\documentclass[letterpaper, 10 pt, conference]{ieeeconf}  % Comment this line out if you need a4paper

\IEEEoverridecommandlockouts                              % This command is only needed if 
\usepackage[T1]{fontenc}
\usepackage{amsmath}
\usepackage{bm}
\usepackage{graphicx}
\usepackage{array}
\usepackage{multirow}
\usepackage{booktabs}
\usepackage{CJKutf8}
\usepackage{subfigure}
\usepackage{color}
\usepackage{amssymb}
\usepackage{cite}
\usepackage{xcolor}
\usepackage{eso-pic}
\usepackage{lipsum}
\usepackage{hyperref}

\title{\LARGE \bf
ICCO: Learning an Instruction-conditioned Coordinator for Language-guided Task-aligned Multi-robot Control
}

\author{Yoshiki Yano$^{1}$, Kazuki Shibata$^{1}$, Maarten Kokshoorn$^{1,2}$ and Takamitsu Matsubara$^{1}$% stops a space
\thanks{$^{1}$ All the authors are with the Division of Information Science, Graduate School of Science and Technology, Nara Institute of Science and Technology (NAIST), Nara, Japan.}
\thanks{$^{2}$ M. Kokshoorn is with the Department of Cognitive Robotics, Faculty of Mechanical Engineering, Delft University of Technology, Delft, Netherlands. }
}

\begin{document}
\iffalse
\AddToShipoutPicture*{%
    \AtPageUpperLeft{%
        \raisebox{-1.2cm}{ % ここで下方向に調整
            \hspace{1cm} % 左側のマージン
            \parbox{\textwidth}{\centering \small 
                This work has been submitted to the IEEE for possible publication. Copyright may be transferred without notice, after which this version may no longer be accessible.
            }
        }
    }
}
\fi

\maketitle
\thispagestyle{empty}
\pagestyle{empty}

%%%%%%%%%%%%%%%%%%%%%%%%%%%%%%%%%%%%%%%%%%%%%%%%%%%%%%%%%%%%%%%%%%%%%%%%%%%%%%%%
\begin{abstract}
Recent advances in Large Language Models (LLMs) have permitted the development of language-guided multi-robot systems, which allow robots to execute tasks based on natural language instructions. However, achieving effective coordination in distributed multi-agent environments remains challenging due to (1) misalignment between instructions and task requirements and (2) inconsistency in robot behaviors when they independently interpret ambiguous instructions. To address these challenges, we propose Instruction-Conditioned Coordinator (ICCO), a Multi-Agent Reinforcement Learning (MARL) framework designed to enhance coordination in language-guided multi-robot systems. ICCO consists of a Coordinator agent and multiple Local Agents, where the Coordinator generates Task-Aligned and Consistent Instructions (TACI) by integrating language instructions with environmental states, ensuring task alignment and behavioral consistency. The Coordinator and Local Agents are jointly trained to optimize a reward function that balances task efficiency and instruction following. A Consistency Enhancement Term is added to the learning objective to maximize mutual information between instructions and robot behaviors, further improving coordination. Simulation and real-world experiments validate the effectiveness of ICCO in achieving language-guided task-aligned multi-robot control. The demonstration can be found at \url{https://yanoyoshiki.github.io/ICCO/}.
\end{abstract}

%%%%%%%%%%%%%%%%%%%%%%%%%%%%%%%%%%%%%%%%%%%%%%%%%%%%%%%%%%%%%%%%%%%%%%%%%%%%%%%%
\section{INTRODUCTION}

With the emergence of Large Language Models (LLMs) \cite{gpt4o}, their language understanding and representation capabilities have attracted attention for application to language-guided robot control. Most studies of language-guided robot control \cite{Ding2023, Chu2024, pang2024kalm} have implicitly assumed that language instructions are consistent with task objectives. However, such assumptions may be problematic in dynamic environments (e.g., manipulation tasks where objects are dynamically relocated or navigation tasks where pedestrians traverse the robot's planned path). Addressing this challenge requires enhanced task-aligned autonomy, allowing robots to interpret and adapt instructions while balancing real-world task requirements and instruction following. This is particularly crucial in distributed multi-robot systems, where multiple robots operate under decentralized control while relying on local observations to execute a number of complex tasks, such as cooperative transportation \cite{Wang, Culbertson, Bernard}, navigation \cite{Zhai, Han, SAMARL, Arul}, and SLAM \cite{multirobot_slam, Chang, Tian}. Accordingly, this study focuses on language-guided multi-robot control in dynamic environments.

Concretely, translating language instructions into coordinated multi-robot behaviors poses two key challenges:
\begin{itemize}
\item [1)] {\it Misalignment} between instructions and task requirements: Language instructions may not fully capture the specific requirements of the given task or may even contradict them, leading to execution discrepancies.

\item [2)]{\it Inconsistency} between instructions and robot behaviors. Ambiguous instructions interpreted solely through the local observations of each robot may lead to conflicting actions among robots, disrupting team coordination.
\end{itemize}

Here, for example, we consider a resource collection task in which multiple robots cooperate to collect resources while defending against invaders (Fig. 1), following a simulation task in a previous work \cite{COPA}. The environment is dynamic because the invaders and resources are randomly located and the invaders move toward the center in launching their invasion. The task requirements consist of two objectives: defending against the invaders and collecting resources. When the instructor gives an abstract instruction such as ``Move to Center,'' robots that do not detect any nearby invaders or resources should follow this instruction immediately. On the other hand, a robot that is close to an invader or a resource is expected to complete its respective task flexibly before following the instruction. However, achieving these objectives is not straightforward. If each robot independently interprets the instruction and task alignment based solely on its own local observations, inconsistencies in behaviors may arise, possibly leading to inefficient coordination and inappropriate task execution.

\begin{figure}[t]
    \centering 
    \includegraphics[width=0.49\textwidth]{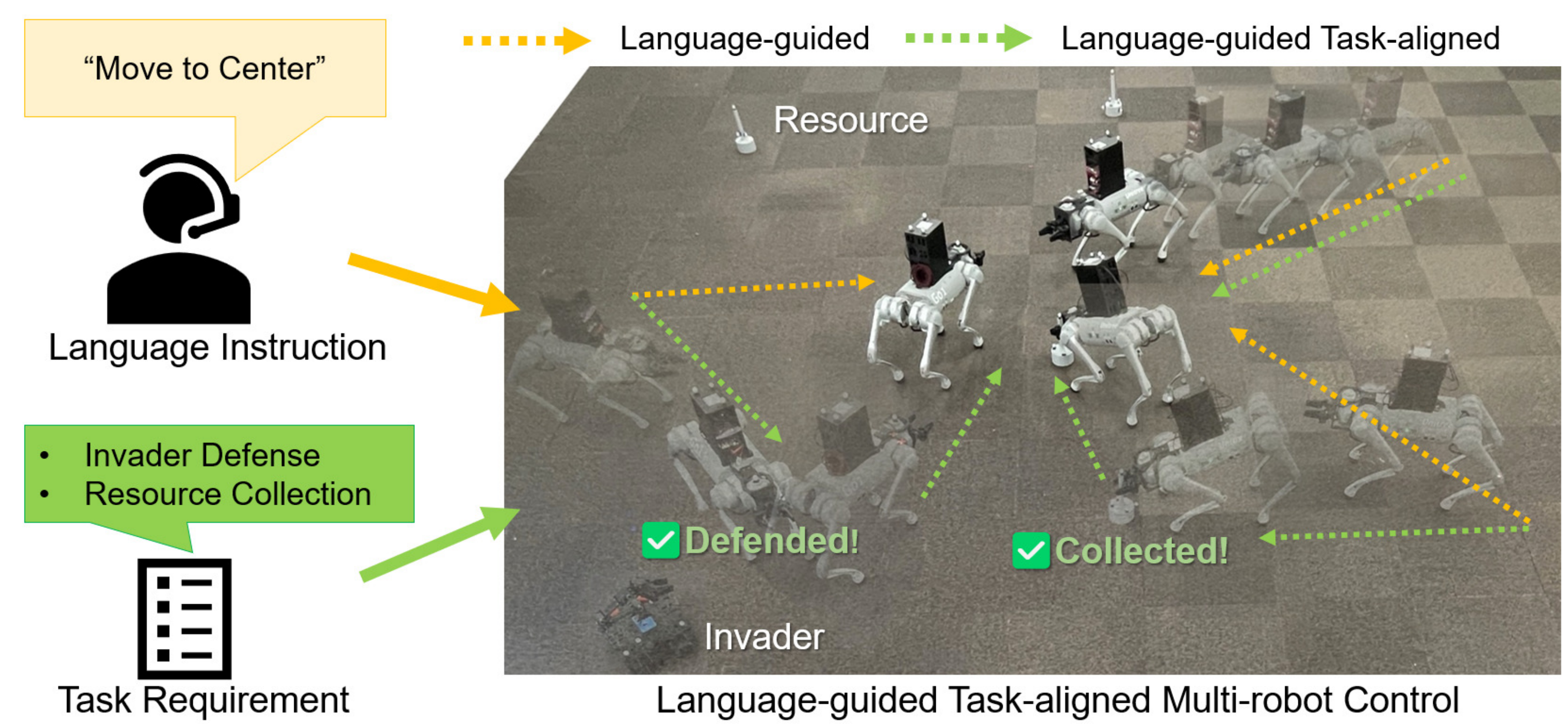}%0.49
    \caption{Overview of language-guided task-aligned multi-robot control in the resource collection task. This figure illustrates how multiple robots coordinate their actions based on language-guided instructions while balancing resource collection and defense against invaders. Robots that are far from invaders or resources follow an instruction (e.g., ``Move to Center'') immediately, while those actively engaged in the task complete their current objective before complying flexibly.}
    \label{fig:framework_overview}
\end{figure}

Prior work on LLM-integrated Multi-Agent Reinforcement Learning (MARL) has explored methods for interpreting and executing language instructions by distributed agents \cite{omron}; however, those methods require extensive inter-robot communication, limiting applicability. Another approach assigns individual instructions to each agent \cite{prorok}, placing a heavy burden on the instructor. To address these issues, this paper explores an alternative framework that enables both language-instruction following and cooperative task execution without relying on inter-agent communication or excessive instructor effort. Since task coordination and consistency require a global perspective, centralized instruction coordination appears to be a reasonable approach.

This paper proposes Instruction-Conditioned Coordinator (ICCO), a MARL framework that balances language-instruction following and cooperative task execution. As shown in Fig. \ref{fig:block_d}, ICCO comprises a Coordinator, which generates and distributes Task-Aligned and Consistent Instructions (TACI) based on language instructions and global environment observations, thus ensuring task alignment and behavioral consistency, and Local Agents, which determine actions using local observations and TACI. ICCO follows the Centralized Training with Decentralized Execution (CTDE) paradigm, jointly optimizing the Coordinator and Local Agent policies to balance instruction following and task requirements. A Consistency Enhancing (CE) Term is introduced in the learning objectives to maximize the lower bound of mutual information between instructions and agent behaviors, improving coordination consistency.

Inspired by a MARL study that enhanced coordination among multiple robots by distributing global messages \cite{COPA}, our approach uses one-way broadcast communication, eliminating inter-agent communication and simplifying system configuration. It also reduces the instructor's burden by removing the need to assign instructions to each agent manually.

The contributions of this study are as follows: 
\begin{itemize} 
    \item We propose ICCO, a MARL framework that balances language-instruction following and cooperative task execution without relying on inter-agent communication or increasing instructor burden.
    \item We validate the effectiveness of the proposed method through simulation experiments in a language-instructed environment.
    \item We confirm the effectiveness of the proposed method by conducting real-world demonstrations using physical robots.
\end{itemize}

\section{RELATED WORKS}

\subsection{Multi-robot control using LLMs}
In recent years, multi-robot control using LLMs has attracted attention in its application to various tasks, including cooperative pick-and-place tasks \cite{roco,merideme}, home service tasks \cite{Smart_llm,S-ATLAS,llmplan}, cooperative navigation \cite{AutoTAMP, Co-NavGPT}, and formation tasks \cite{omron,prorok}. Furthermore, several recent studies \cite{roco,S-ATLAS,merideme} have proposed decentralized approaches in which each agent is equipped with its own LLM and controls itself based on task-planning via interactive dialogues.

However, dialogue-based approaches \cite{roco,S-ATLAS,merideme} suffer from increased LLM inference times as tasks progress, leading to prolonged decision-making latency. In contrast, our study employs MARL policies for agent decision-making to prevent triggering of LLM inference during action selection, which permits multi-robot control without delays.

\subsection{Multi-robot control using LLMs and MARL}
Several recent studies \cite{prorok,omron} have integrated LLMs with MARL frameworks \cite{IQL,MADDPG} to achieve both language-instruction following and successful task execution. Liu et al. \cite{omron} proposed a MARL approach that achieves compliance with language instructions and task execution through inter-agent communication. However, this method necessitates extensive communication among agents, thus limiting its applicability. Morad et al. \cite{prorok} introduced another MARL approach in which an instructor provides specific instructions to each agent individually. However, this approach is impractical for real-world applications due to the high cognitive burden placed on the instructor.

This study differs from those approaches \cite{prorok,omron} in that it introduces a Coordinator that concurrently facilitates both language instructions and coordinated behavior. In addition, it incorporates the mutual information between instructions and agent behaviors, thus improving coordination consistency. Consequently, our method can balance instruction following and cooperative task execution without requiring inter-agent communication or individual instructions given to each agent.

    \section{PRELIMINARY}
        \subsection{Decentralized Partially Observable Markov Decision Process (Dec-POMDP)}
           This study formulates the multi-robot control problem as a Dec-POMDP \cite{DecPOMDP}. We define the tuple $\langle\mathcal{S}, \mathcal{B}, \{\mathcal{A}^i\}^{i\in \mathcal{B}}, P, r, \{\mathcal{O}^i\}^{i\in \mathcal{B}}, \gamma\rangle$, where $s\in \mathcal{S}$ denotes the global state of the environment. At each time step $t$, each agent $i \in \mathcal{B}:=\{1,\cdots,n\}$ receives a local observation $o^i\in \mathcal{O}^i$ and selects an action $a^i \in \mathcal{A}^i$. This yields a joint action $\boldsymbol{a} \in \mathcal{A} = \prod_{i=1}^{n} \mathcal{A}^i$. Subsequently, the environment transitions from the current state $s$ to a new state $s'$ according to the state transition function $P(s'|s,\boldsymbol{a}): \mathcal{S}\times \mathcal{A}\times \mathcal{S}\to [0,1]$, providing the agents with a global reward $r(s,\boldsymbol{a})$. Given a joint policy $\boldsymbol{\pi}:=(\pi^i)^{i\in \mathcal{B}}$, the joint action-value function at time $t$ is defined as $Q^{\boldsymbol{\pi}}(s_t,\boldsymbol{a}_t)=\mathbb{E}\left[R_t|s_t,\boldsymbol{a}_t\right]$, where $R_t = \mathbb{E}\left[\sum_{k=0}^{\infty} \gamma^k r_{t+k}\right]$ denotes the discounted cumulative reward and $0\le \gamma<1$ is a discount factor. The objective of this study is to find a policy that achieves the optimal value $Q^*=\max_{\boldsymbol{\pi}} Q^{\boldsymbol{\pi}}(s_t,\boldsymbol{a}_t)$.

        \subsection{QMIX}
            QMIX \cite{QMIX} is one of the typical MARL algorithms that adopt the CTDE paradigm. During training, it leverages the global state to learn the joint action-value function; during execution, each agent selects actions based on its own policy, using only local observations. QMIX employs a factorization approach, where the joint action-value function $Q^{{\rm tot}}$ is decomposed as a combination of the local action-value functions of individual agents, given by
            \begin{equation}
                Q^{{\rm tot}}_\theta(s,\boldsymbol{a})=f(Q^1(o^1,a^1),...,Q^n(o^n,a^n);s),
                \label{eq: mixing}
            \end{equation}
            where $\frac{\partial{f}}{\partial{Q}^i}\ge 0$, $\forall i\in \mathcal{B}$, and $\theta$ is the weight parameter of the $Q^{\rm tot}$ network.

            The mixing network $f$ is a monotonically increasing function that takes the action values of individual agents as inputs. Consequently, maximizing each agent's local action value leads to the maximization of the joint action value. The joint action-value function is trained based on the Bellman equation $Q^{\rm tot}(s_t, \boldsymbol{a}_t) = \mathbb{E}\Big[r_t + \gamma \operatorname{max}_{\boldsymbol{a}^{\prime}}Q^{\rm tot}(s', \boldsymbol{a}^{\prime})\Big]$. To optimize the action-value function, the loss function
            \begin{equation}
               \begin{aligned}
                   \mathcal{L_\mathrm{RL}}(\theta) = \mathbb{E}_{\left(\boldsymbol{\tau}_t, \boldsymbol{a}_t, r_t\right) \sim \mathcal{D}}\Big[\big(&r_t+ \\
                   \gamma \max_{\boldsymbol{a}^{\prime}} Q_{\theta^{-}}^{\rm tot}\left(\boldsymbol{\tau}_{t+1}, \boldsymbol{a}^{\prime}\right) 
                   &-Q_{\theta}^{\rm tot}(\boldsymbol{\tau}_t, \boldsymbol{a}_t, s_t)\big)^2\Big]
               \end{aligned}
               \label{eq: qmix_update}
            \end{equation}
            is minimized, where $\theta^-$ represents the weight parameters of the target network for $Q^{\rm tot}$. Additionally, the trajectory set is defined as $\boldsymbol{\tau}=\left\{\tau^i \mid i \in \mathcal{B}\right\}$, where each agent's trajectory is given by $\tau^i=\left(o_0^i, a_0^i, \ldots o_t^i\right)$.

    \section{Proposed Method}
        \subsection{Overview of ICCO framework}
      
\begin{figure}[t]
                \begin{center}
                    \includegraphics[width=0.42\textwidth]{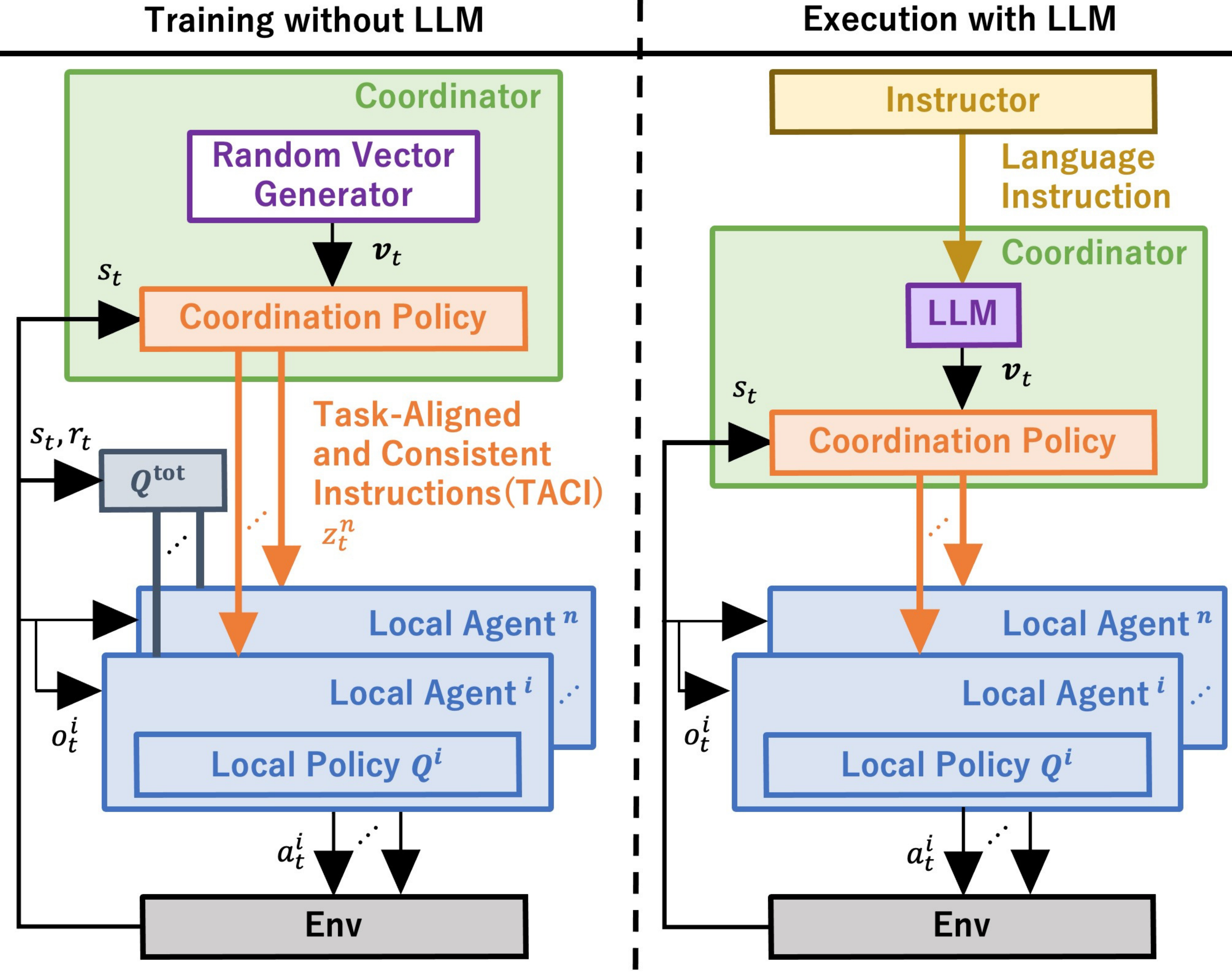}%fig2.pdf}
                    \caption{Block diagram illustrating ICCO framework: Instruction-Conditioned Coordinator using Language Instruction}
                    \label{fig:block_d}
                \end{center}
            \end{figure}

            In this section, we describe ICCO, a MARL framework designed to balance language-instruction following with task execution. An overview of the proposed framework is presented in Fig. \ref{fig:block_d}. ICCO primarily comprises a \textit{Coordinator} and a group of \textit{Local Agents}. 
             
            The Coordinator employs an LLM to convert language instructions from the Instructor into agent-specific vector representations. Subsequently, the coordination policy integrates these vectors with the global state $s_t$ to generate Task-Aligned and Consistent Instructions (TACI) for each agent, denoted by $\boldsymbol{z}_t:=(z_t^1,\cdots,z_t^n)$. Agent $i$ selects its action based on the local observation and the instruction according to its local policy. The coordination policy and local policies are consistently trained within the CTDE framework to maximize a reward function that balances instruction following and task achievement. 
           
            Furthermore, to efficiently learn policies while accommodating diverse language instructions, we adopt the Training without LLM and Execution with LLM approach, where policies are trained without an LLM, which is used only during execution.
    
        \subsection{Policy model}
       
            MARL using LLMs \cite{omron} leverages inter-agent communication and adopts a distributed policy under the local observation. However, in the absence of inter-agent communication, it is challenging to achieve both compliance with language instructions and successful task execution using only local policies. 
            
            To address this issue, ICCO introduces a coordination policy that computes TACI $\boldsymbol{z}_t$ given by
            \begin{align}
                \{ \boldsymbol{\mu}_t, \boldsymbol{\Sigma}_t \} &= g_{\phi}(s_t, \boldsymbol{v}_t), \\
                \boldsymbol{z}_t &\sim \mathcal{N}(\boldsymbol{\mu}_t, \boldsymbol{\Sigma}_t),
            \end{align}
            where $\phi$ represents the parameters of the function $g$, $\mathcal{N}$ denotes a normal distribution, and $\boldsymbol{\mu}$ and $\boldsymbol{\Sigma}$ are the mean and covariance, respectively. 
            
            Furthermore, the local agent has one local policy that computes the action by 
            $a^i_t=\operatorname{argmax}_{a^i_t} Q^i(\tau^i_t, a_t^i | z^i_t),$ where the action $a^i_t$ is selected to maximize the local action-value function $Q^i$ based on the local observation $o^i_t$ and individual TACI $z^i_t$. Following QMIX, this function is modeled by the joint action-value function $Q^{\rm tot}_\theta$ as in Eq. (\ref{eq: mixing}).

        \subsection{Training objectives}
       
            The training objective is to ensure that agents follow language instructions while accomplishing the task. When instructions align with task requirements, the objective function is maximized to train agents to follow the instructions. However, when misalignment occurs, the instructions must be adjusted to balance instruction following and task execution. Additionally, due to partial observability, local agents may select different actions even when given the same instructions, necessitating better consistency in robot behavior. To address these challenges, the objective function promotes both the alignment between task requirements and instructions and the consistency of multi-robot behaviors.

            {\bf Task-Alignment Term:}
            To promote task alignment, we introduce task achievement and instruction following in the reward function, formulating the task-alignment term using (\ref{eq: qmix_update}) as follows:
                \begin{equation}
                    \begin{aligned}
                    & \mathcal{L}_{\mathrm{RL}}(\theta, \phi)=\mathbb{E}_{\left( \boldsymbol{\tau}_t, \boldsymbol{a}_t, r_t, s_{t}, s'\right) \sim \mathcal{D}}\left[\left(r_t+\right.\right. \\
                    & \left.\left.\gamma \max _{\boldsymbol{a}^{\prime}} Q_{\theta^{-}}^{\mathrm{tot}}\left(\boldsymbol{\tau}_{t+1}, \boldsymbol{a}^{\prime} \mid \boldsymbol{z}_{t+1}\right)-Q_\theta^{\mathrm{tot}}\left(\boldsymbol{\tau}_t, \boldsymbol{a}_t \mid \boldsymbol{z}_{t} \right)\right)^2\right],
                    \end{aligned}
                    \label{eq: rl loss}
                \end{equation}
            where $Q_\theta^{\mathrm{tot}}$ and $Q_{\theta^{-}}^{\mathrm{tot}}$ are conditioned on $\boldsymbol{z}_t$ and $\boldsymbol{z}_{t+1}$, respectively.
            
            {\bf Consistency Enhancing Term:}
            To promote consistency between instructions and multi-robot behaviors, we consider the mutual information term, given by
            \begin{equation}
            I\left(\boldsymbol{z}_t ; \boldsymbol{\zeta}_t, s_t, \boldsymbol{v}_t\right) = D_{\mathrm{KL}} \left( p(\boldsymbol{z}_t \mid \boldsymbol{\zeta}_t, s_t, \boldsymbol{v}_t) \, \big\| \, p(\boldsymbol{z}_t \mid s_t, \boldsymbol{v}_t) \right),
            \end{equation}
            where \( \boldsymbol{\zeta}_t = \left(\boldsymbol{o}_{t+1}, \boldsymbol{a}_{t+1}, \ldots, \boldsymbol{o}_{t+T-1}, \boldsymbol{a}_{t+T-1} \right) \) represents the observation-action trajectory from step \( t+1 \) to \( t+T-1 \).
            
            Since the future observation-action trajectory \( \boldsymbol{\zeta}_t \) is high-dimensional and varies stochastically, computing the posterior distribution \( p(\boldsymbol{z}_t \mid \boldsymbol{\zeta}_t, s_t, \boldsymbol{v}_t) \) is intractable. To address this, we introduce an approximate posterior distribution \( q_{\xi}(\boldsymbol{z}_t \mid \boldsymbol{\zeta}_t, s_t, \boldsymbol{v}_t) \) in a factorized form \cite{Rakelly} as follows:
            \begin{equation}
                \begin{split}
                    q_{\xi}\left(\boldsymbol{z}_t \mid \boldsymbol{\zeta}_t, s_t, \boldsymbol{v}_t\right) \propto & \,
                    q_{\xi}^{(t)}\left(\boldsymbol{z}_t \mid s_t, \boldsymbol{a}_t, \boldsymbol{v}_t\right) \\
                    & \prod_{k=t+1}^{t + T - 1} q_{\xi}^{(k)}\left(\boldsymbol{z}_t \mid \boldsymbol{o}_k, \boldsymbol{a}_k, \boldsymbol{v}_k\right).
                \end{split}
                \label{eq: q distribution}
            \end{equation}

            Using (\ref{eq: q distribution}), the lower bound of \( I\left(\boldsymbol{z}_t ; \boldsymbol{\zeta}_t, s_t, \boldsymbol{v}_t\right) \) can be derived using variational inference \cite{COPA} as follows:
            \begin{equation}
                \begin{aligned}
                I\left(\boldsymbol{z}_t ; \boldsymbol{\zeta}_t, s_t, \boldsymbol{v}_t\right) \geq 
                \mathbb{E}_{s_t, \boldsymbol{z}_t, \boldsymbol{\zeta}_{t},\boldsymbol{v}_t} \left[\log q_{\xi}\left(\boldsymbol{z}_t
                \mid \boldsymbol{\zeta}_t, s_t, \boldsymbol{v}_t\right)\right]
                \\ 
                + H\left(\boldsymbol{z}_t \mid s_t, \boldsymbol{v}_t\right),
                \end{aligned}
                \label{eq: mutual information}
            \end{equation} 
            where \( H \) denotes entropy. Using (\ref{eq: mutual information}), we introduce the consistency-enhancing term as follows:
            \begin{equation}
                \begin{aligned}
                    \mathcal{L}_{\mathrm{CE}}(\phi, \xi) = &  
                    -\mathbb{E}_{s_t, \boldsymbol{z}_t, \boldsymbol{\zeta}_t, \boldsymbol{v}_t} 
                    \Big[ \log q_{\xi} \left(\boldsymbol{z}_t \mid \boldsymbol{\zeta}_t, s_t, \boldsymbol{v}_t\right) \Big] \\
                    & - H\left(\boldsymbol{z}_t \mid s_t, \boldsymbol{v}_t\right).
                \end{aligned}
                \label{eq: ce loss}
            \end{equation}
            
            During the training phase, the network parameters \( (\theta, \phi, \xi) \) are updated by minimizing $\mathcal{L}_{\mathrm{RL}} + \mathcal{L}_{\mathrm{CE}}$.
            
        \subsection{Training without LLM and execution with LLM}
            
            In MARL with LLMs \cite{prorok}, using training policies for diverse language expressions that are not intrinsic to the task results in extended training durations. Moreover, using LLMs during training necessitates repeated inference within the training loop, leading to a substantial increase in computational cost.

            To address this issue, we adopt a training approach where policies are learned without LLMs during training but then executed with LLMs, which are adopted only during this latter phase following the method proposed previously \cite{omron}. Specifically, during the training phase, policies are trained by random sampling within the expected instruction vector space without using an LLM. Each instruction vector is generated as a smooth trajectory by sampling Gaussian noise $\boldsymbol{\epsilon}_k$ at each time step $k$ and computing the cumulative sum from the agent's current position, given by $\boldsymbol{v}_{t+k}=\boldsymbol{p}_t+\sum^K_{k=1}\boldsymbol{\epsilon}_k$. During the execution phase, the LLM converts diverse language instructions into instruction vectors for each agent. These vectors are clipped within a predefined value to ensure that agents stay within the movable region, and to prevent the LLM outputs from becoming out-of-distribution relative to the training data. This approach reduces training time, compared to methods that incorporate LLMs during training, while still maintaining the capability to handle various language instructions.
                    
    \section{EXPERIMENT}
    
        In this section, we validate the effectiveness of the proposed method through simulations and multi-robot experiments that involve both instruction following and task execution. The key research questions in this experiment are as follows:

        \begin{itemize}
            \item Can the policy learned by ICCO balance instruction following and task execution? (\ref{sec: 4area_test})
            \item Can ICCO perform effectively with natural language instructions using an LLM? (V-C)
            \item Can the Coordinator in ICCO be replaced by an LLM with task-aligned prompt engineering? (\ref{sec: ex_coordinator_policy})
        \end{itemize}

       In the following experiments, training and evaluation were conducted in simulation, while a demonstration of the trained policies was performed in a real environment.
         
         \subsection{Experimental setup}
            \subsubsection{\bf{Simulation Setup}}
        
\begin{figure}[t]
                    \begin{center}
                        \includegraphics[width= 70mm]{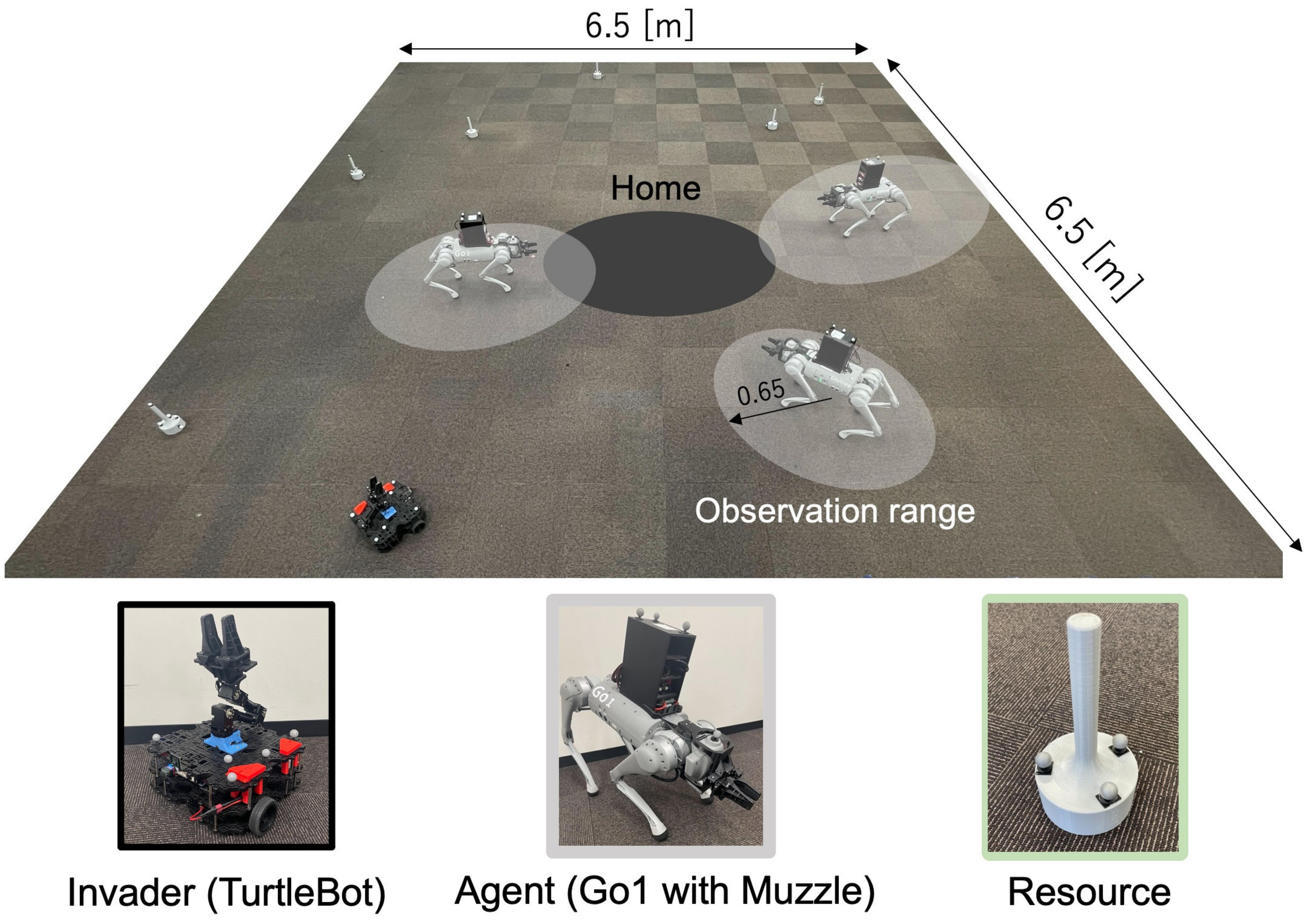}%{figures/real_env.pdf}%{figures/fig3.pdf}
                        \caption{Resource Collection Environment}
                        \label{fig: Sim_view}
                    \end{center}
                \end{figure}

              The simulation environment used in this study is based on the resource collection environment from \cite{COPA}, but it is adjusted to our real environment, as shown in Fig. \ref{fig: Sim_view}. The environment contains three homogeneous agents, an invader, and six resources, whose initial positions are randomly generated within a $6.5\times 6.5$ m area. Note that when an agent picks up a resource, a new resource is spawned. An agent can transport a resource upon contact. The task objective is that Agents collect resources and transport them to the home base while preventing an invader from reaching it.
              
                Each agent can observe only other agents, resources, the invader and the home within a circle of radius 0.65 m. The instruction vector $\boldsymbol{v}_t$ consists of the instruction vectors for all agents, where each agent's instruction is specified by waypoints. The global state provided to the coordination policy includes the positions and velocities of all agents, binary flags indicating whether they are carrying resources or defending invaders, the positions of environmental elements such as resources and invaders, and instruction vectors generated by the LLM for all agents.
                
                The agent's reward is given by $r=r_{\rm task}+r_{\rm inst}$, where $r_{\rm task}$ and $r_{\rm inst}$ represent the task and instruction-following rewards, respectively. 
                The task reward is given by $r_{\rm task} = r_{\rm pick} + r_{\rm collect} + r_{\rm defense}$, where $r_{\rm pick} = 5$ for picking a resource, $ r_{\rm collect} = 1$ for releasing it at home, $ r_{\rm defense} = 4$ for hitting an invader, and $ r_{\rm defense} = -4$ if the invader reaches home. The instruction-following reward is given by $r_{\rm inst} = 1.3(e_{\mathrm{cossim}} + 0.1e_{\mathrm{dist}})$, where \( e_{\mathrm{cossim}} \) is the cosine similarity between two displacement vectors: one from the agent's current position to the next position and one from the nearest waypoint to the second-nearest waypoint, and \( e_{\mathrm{dist}} \) is the distance between the agent's current position and the nearest waypoint.

            \subsubsection{\bf{Prompt Setup}}

                The key components of the prompt are shown in Fig. \ref{fig:prompt}. To configure the spatial setting, the LLM is anchored to a 2D field. Empirical observations indicate that this enables the LLM to interpret ``up'' as the positive $y$-axis direction and ``right'' as the positive $x$-axis direction without requiring additional specifications in the prompt. The inputs to the LLM consist of language instructions and initial configurations, while the output comprises waypoints for each agent. In the instruction part, the phrase ``STRICTLY COMPLY with the following order'' ensures strict following of the given instruction. Furthermore, to generate stable trajectories, we adopt a few-step instruction approach following previous works \cite{chain,reason}, where the trajectory is generated in two stages. Specifically, in the Movement Strategy stage, the LLM explains the trajectory based on the given instructions, and in the Trajectory Generation stage, it outputs the waypoints for each agent. Since this study focuses on handling task-misaligned language instructions, task-related information is omitted from the prompt for simplicity. The experiment was conducted by feeding these prompts into GPT 4o \cite{gpt4oAPI}.

                \iffalse
                \begin{figure}[H]
                    \vspace{-10pt}
                    \includegraphics[width=0.45\textwidth]{figures/LLM_role.pdf}
                    \vspace{-10pt}
                \end{figure}
                \fi

                \iffalse
                \begin{figure}[H]
                    \vspace{-10pt}
                    \includegraphics[width=0.45\textwidth]{figures/few_inst.pdf}
                    \vspace{-10pt}
                \end{figure}
                \fi

            \begin{figure}[t]
                \begin{center}
                    \includegraphics[width=0.4\textwidth]{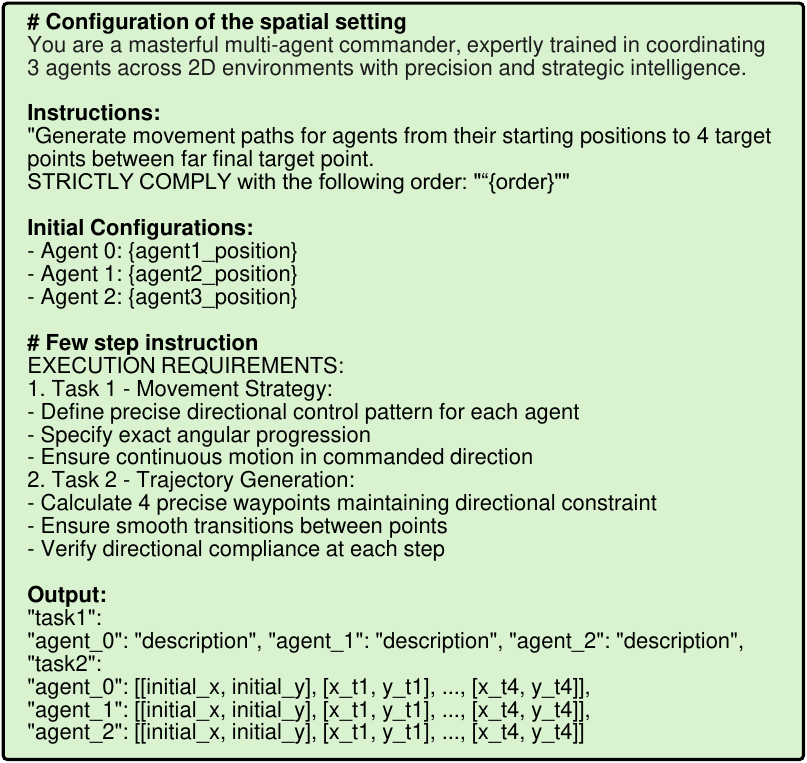}
                    \caption{Key components of the prompt for the ICCO's LLM: Elements enclosed in \{\} represent variables that are updated when inputted to the LLM.}
                    \label{fig:prompt}
                \end{center}
            \end{figure}

            \subsubsection{\bf{Baselines}}
           
             To evaluate the effectiveness of the proposed method ({\bf ICCO}), comparisons were made with the following baseline approaches.
            %\begin{itemize}
            %    \item QMIX\cite{QMIX}: Coordinatorの有効性を確認するため、言語指示と各エージェントの観測の下での方策を採用したQMIXベースの手法と比較した。この手法は、提案手法と同様に、エージェント間の通信を用いない。
            %    \item QMIX (FULL): Coordinatorの有効性を確認するため、言語指示と全観測の下での方策を採用したQMIXベースの手法と比較した。この手法は、全てのエージェントと通信できる大域通信を採用した。
            %    \item ICCO w/o CE: 提案法のConsistency Enhancing Termの有効性を確認するため、式(\ref{eq: ce loss})の$\mathcal{L}_{\mathrm{CE}}(\phi, \xi)$なしで、方策を訓練するICCOと比較した。
            %\end{itemize}
            \begin{itemize}
                \item {\bf QMIX} \cite{QMIX}: To evaluate the effectiveness of the coordination policy, we compared our approach with a QMIX-based method using policies conditioned on language instructions and individual agent observations, without inter-agent communication.
                \item {\bf QMIX (FULL)}: Similar to {\bf QMIX}, but this method allows each agent to access the global state.
                \item {\bf ICCO w/o CE}: To confirm the effectiveness of the CE term in the proposed method, we conducted an ablation study comparing {\bf ICCO} without the $\mathcal{L}_{\mathrm{CE}}(\phi, \xi)$ used in Eq. (\ref{eq: ce loss}).
            \end{itemize}

            {\bf QMIX} and {\bf ICCO} adopt the same architecture as \cite{QMIX} and \cite{COPA}, except for differences in input dimensionality resulting from the presence of instruction vectors.

           All methods were trained in the following settings: Agents, Invaders, and Resources are sampled uniformly in the field's range for each episode, which consists of 145 steps. Instruction vectors are generated every four steps using a zero-mean Gaussian distribution from the agent's current position to promote a smooth trajectory. 

            \subsubsection{\bf{Real robot system}}
           
                We demonstrated use of a real robot Go1-M, which is a Unitree Go1 with a muzzle based on the OpenManipulator-X gripper. A motion capture system observed the positions of Go1-M, the resources, and the invader. Due to differences in locomotion mechanisms between Go1-M and the simulation agents, real-world transitions were computed from policy outputs, with position control using the Dynamic Window Approach (DWA) \cite{DWA}. A pre-designed controller was used for the catch-and-release action required for resource picking and collection.

        \subsection{Evaluation of instruction following and task execution}\label{sec: 4area_test}              
          
           The performances of the policies learned by all methods were evaluated in terms of reward values. For the evaluation, the field was divided into four quadrants. Within each quadrant, 145 steps were processed using a smooth instruction vector within the quadrant, generated using a zero-mean Gaussian distribution from the current agent position. The target quadrants were sequentially transitioned from the first to the fourth quadrant, resulting in 580 steps per episode. Each evaluation consisted of 20 trials.

\begin{figure}[t]
                \begin{center}
                    \includegraphics[width=0.4\textwidth]{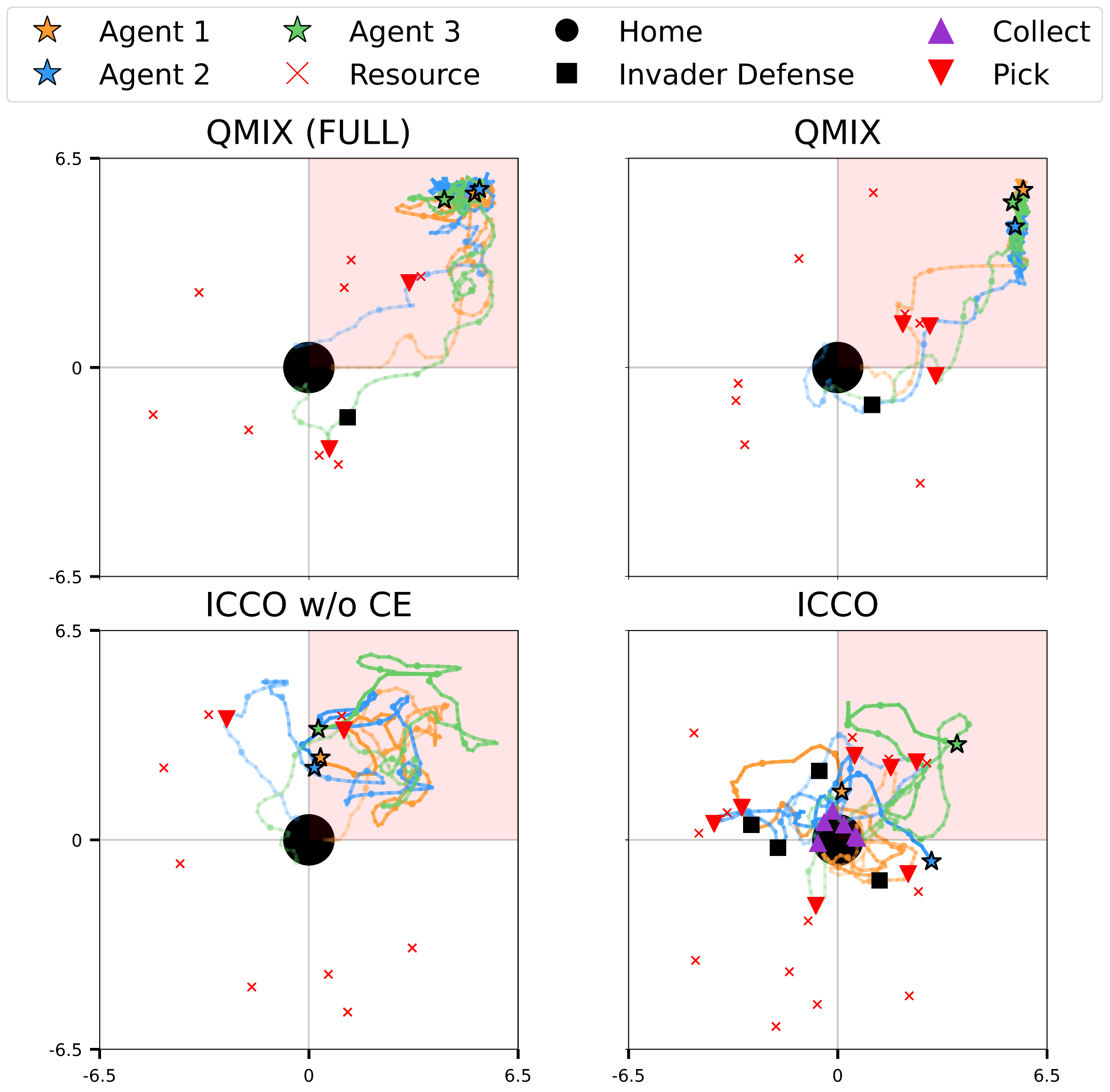}%{figures/fig4.pdf}
                    \caption{Agent trajectories with instruction vectors generated in the red-shaded quadrant. ICCO is the only approach that successfully achieved resource collection and invader defense while approximately controlling agents in the designated quadrant, whereas others failed to perform these tasks.}
                    \label{fig:4area_trj}
                \end{center}
            \end{figure}

            Fig. \ref{fig:4area_trj} shows the detailed results for the first quadrant with all four methods. In {\bf QMIX} and {\bf QMIX (FULL)}, agents moved within the designated quadrant but failed to transport resources to home or defend against invaders. Similarly, {\bf ICCO w/o CE} did not effectively achieve resource collection or defense against invaders. In contrast, {\bf ICCO} approximately controlled agents within the quadrant while enabling them to achieve both resource collection and invader defense, indicating that the CE term improved alignment between instruction following and agent behavior. Consequently, {\bf ICCO} achieved the highest reward among the evaluated methods. The rewards of the learned policies are summarized in Fig. \ref{fig: 4area_reward_bar}.
            
            Overall, the proposed method demonstrated superior instruction following and task execution compared to the baseline methods.

        \subsection{Effectiveness of ICCO with natural language instructions and LLM}\label{sec: ex_LLM_trj}
        
            We validated the proposed method by generating instruction vectors using an LLM. Fig. \ref{fig: real_strobe} shows a demonstration from the simulation experiments in Subsection \ref{sec: 4area_test}, where the trained policy was tested with different language instructions as listed in Table \ref{tab:agent_instructions}.

\begin{figure}[t]
                \centering
                \subfigure[]{
                    \includegraphics[width=0.2\textwidth]{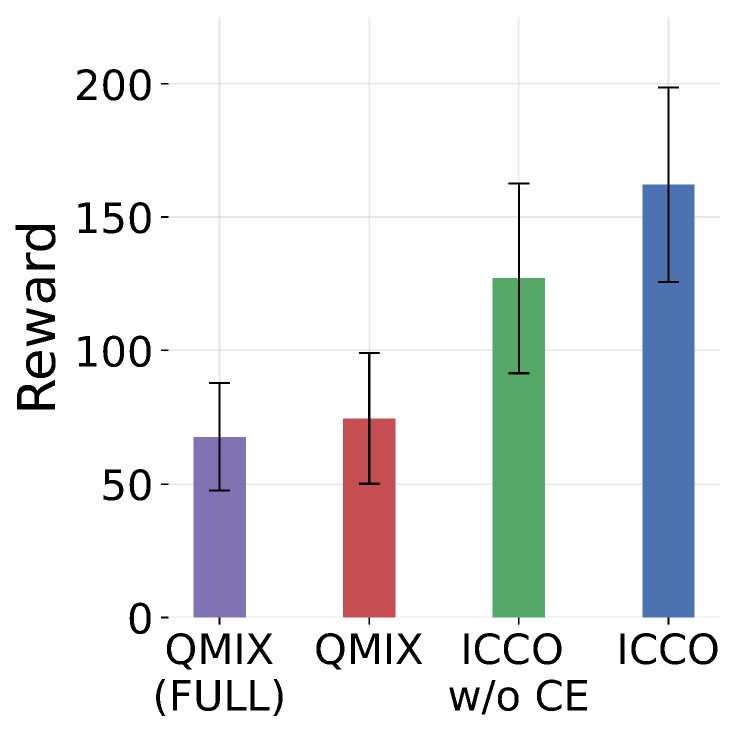}
                    \label{fig: 4area_reward_bar}
                }
                \subfigure[]{
                    \includegraphics[width=0.2\textwidth]{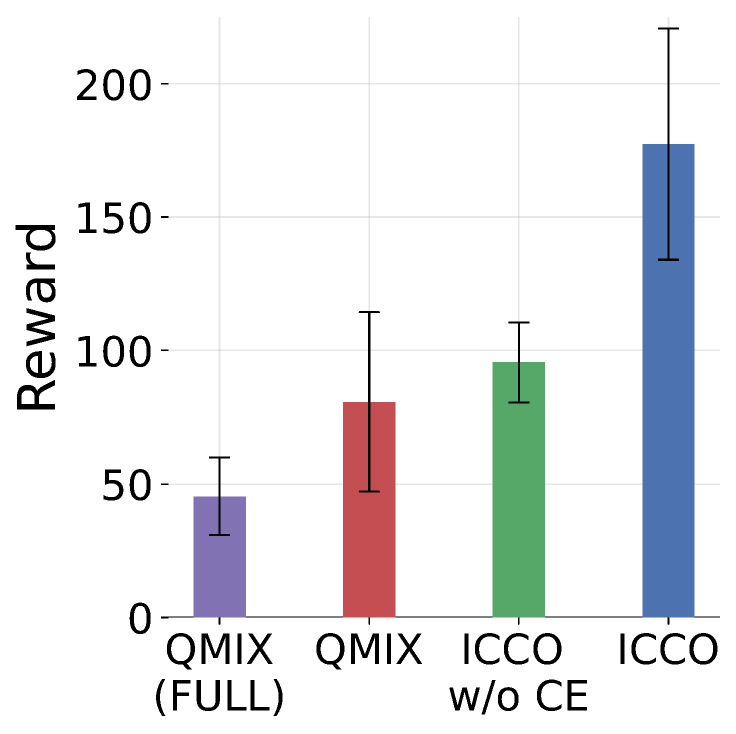}
                    \label{fig: llm_reward_bar}
                }
              
                \caption{Comparison of rewards: (a) rewards from an experiment using reference trajectories as instructions every 4 steps in four quadrants; (b) rewards from an experiment using four natural language instructions via LLMs, given every 145 steps. Each time a language instruction was given, the prompt was re-fed to the LLM. Each experiment was conducted for 145 continuous steps per quadrant and language instruction.}
                \label{fig: reward_bar}
            \end{figure}

\begin{figure*}[t]
                \begin{center}
                    \includegraphics[width=0.7\textwidth]{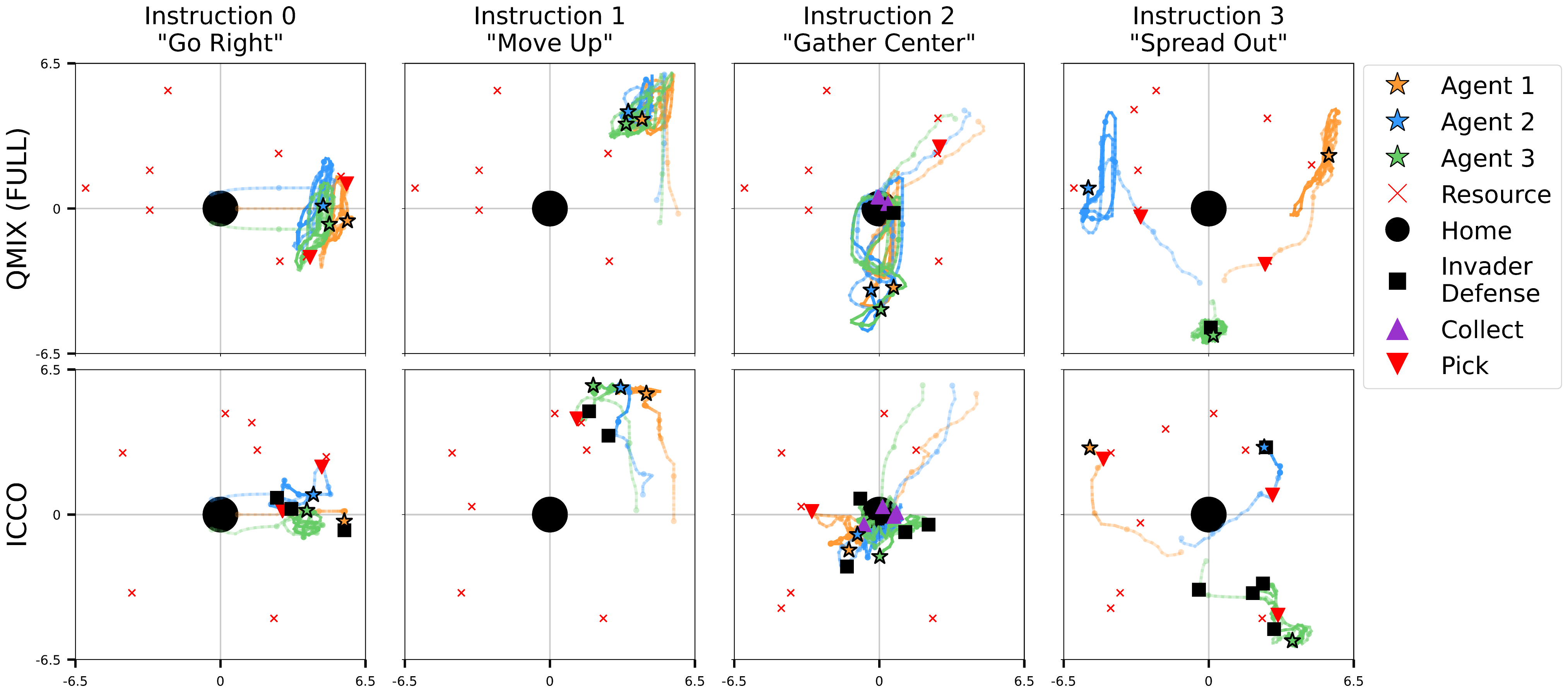}
                    \caption{Agent trajectories for four different language instructions. In ``Gather Center,'' where instruction following aligns with task execution, {\bf ICCO} outperformed {\bf QMIX (FULL)} in gathering agents, collecting resources, and defending against invaders. In other tasks, {\bf QMIX (FULL)} followed instructions but failed in resource collection and defense, while {\bf ICCO} was less effective in resource collection but successfully defended against invaders.}
                    \label{fig: real_strobe}
                \end{center}
            \end{figure*}

\begin{figure*}[t]
            \vspace{-5pt}
                \centering
                \subfigure[{\bf QMIX-FULL}]{
                    \includegraphics[width=0.77\textwidth]{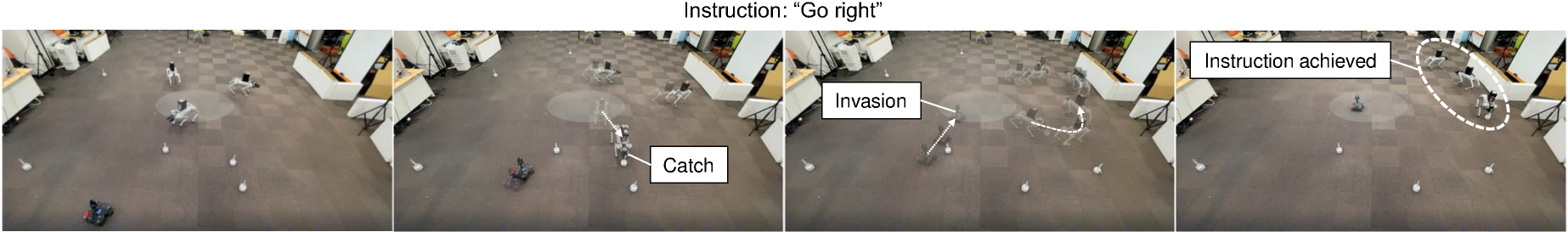}
                    \label{fig: qmixfull_demo}
                }
                \subfigure[{\bf ICCO}]{
                    \includegraphics[width=0.77\textwidth]{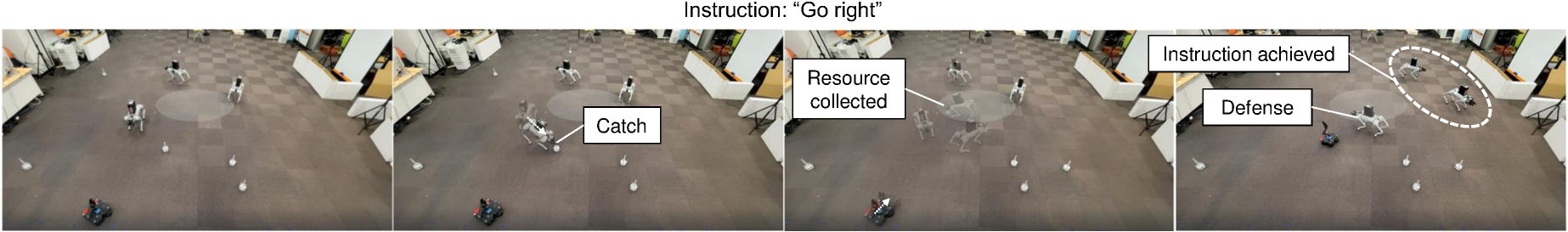}
                    \label{fig: icco_demo}
                }
                \caption{Real-robot demonstrations}
                \label{fig: demo}
            \end{figure*}

            In the ``Gather Center'' task, where instruction following aligns with task execution, {\bf ICCO} outperformed {\bf QMIX (FULL)} by gathering agents closer to the center while collecting more resources and defending against more invaders. In the remaining tasks, where misalignment exists between instruction following and task execution, {\bf QMIX (FULL)} successfully followed the instructions but largely failed in resource collection and invader defense. In contrast, {\bf ICCO} was less effective in resource collection but successfully defended against invaders. Fig. \ref{fig: llm_reward_bar} shows the rewards obtained over 20 trials for each method across four different instructions. The results indicate that {\bf ICCO} achieved higher rewards than the baseline methods. \textcolor{black}{Additionally, Fig. \ref{fig: demo} presents real-robot demonstrations using the trained policies, where the language instruction ``Go right'' was provided. The results of the real-robot experiments show similar outcomes to those observed in the simulation.} See the attached video for more details on the real-robot experiments.

            In summary, the proposed method can effectively operate with natural language instructions using an LLM.

            \begin{table}[t]
                \centering
                \vspace{-10pt}
                \caption{Language instructions}
                \begin{tabular}{p{1.8cm} p{6cm}}
                    \toprule
                    \textbf{Instruction} & \textbf{Instruction Details} \\
                    \midrule
                    Go Right & The agents need to form a line formation on the right side. \\
                    Move Top & The agents need to form a line formation at the top. \\
                    Gather Center & The agents need to gather at the (0,0) position. \\
                    Spread Out & The agents must spread out from the center. \\
                    \bottomrule
                \end{tabular}
                \label{tab:agent_instructions}
            \end{table}

        \subsection{Can the Coordinator in ICCO be replaced by an LLM with task-aligned prompt engineering?}
        \label{sec: ex_coordinator_policy}
        
            {\bf Setup:} In this experiment, we further investigated the effectiveness of ICCO by comparing it to an approach that employs task-aligned prompts as a complementary method instead of using a Coordinator. Since the previous two experiments confirmed instruction following, this experiment focused on evaluating task performance. We compared our method with the following methods:
            \begin{itemize}
                \item {\bf QMIX with Task-Aligned Prompt (QMIX-TAP)}: A method that combines {\bf QMIX} with an LLM using task-aligned prompts.
                \item {\bf QMIX-TAP without task reward (QMIX-TAP w/o TR)}: A method based on QMIX-TAP, excluding the task reward (TR).
            \end{itemize}

            These methods commonly used a task-aligned prompt that includes resources, invaders, and task rewards. See Appendix for further details on the prompt.

                % \greenboxtext{Prompt: Omni LLM \
                % ------------------------------------------------------------------ \
                % Resource collection task notation: (Detail of task defenition)\
                % Reward Setting: (Detail of reward setting)\
                % Each compornen: [[agent1 xy],..,[agent3xy],[resource1 xy],...,[resource6 xy],[invader xy]]
                % }
                
            % 結果
         
            {\bf Results:} 
            Table \ref{tab:comparison} presents a comparison of the number of resource picks, collections, and invader defenses. {\bf QMIX-TAP} outperformed {\bf QMIX-TAP w/o TR} across all metrics. This result suggests that even when task rewards are provided to the LLM, incorporating task rewards into local RL agents further contributes to task performance. Furthermore, {\bf ICCO} outperformed all other methods in every metric, suggesting that the Coordinator was more effective than an LLM with task-aligned prompt engineering.

            In summary, the Coordinator in ICCO cannot be replaced by an LLM with task-aligned prompt engineering.

        \begin{table}[t] 
          \centering
          \caption{Comparison of the numbers of picks, collections, and invader defenses over 20 trials}
          \begin{tabular}{lccc}
            \toprule
            \textbf{Method} & \textbf{Pick} & \textbf{Collect} & \textbf{Defense} \\
            \midrule
            {\bf QMIX-TAP} & $5.1\pm1.5$ & $3.0\pm1.6$ & $6.8\pm3.2$ \\
            {\bf QMIX-TAP w/o TR} & $3.0\pm1.1$ & $1.1\pm0.7$ & $2.0\pm1.7$ \\
            {\bf ICCO} & ${\bf5.6\pm 2.2}$ & ${\bf3.7\pm 1.8}$ & ${\bf 10.5\pm 2.9}$ \\
            \bottomrule
          \end{tabular}
          \label{tab:comparison}
        \end{table}
            
    \section{DISCUSSION}
    In this study, we conducted a real-robot demonstration on a simple task of resource collection. However, when applied to more complex tasks, such as cooperative transport, a policy trained in simulation may perform worse in the real world. To address this issue, the robustness of the trained policy should be enhanced using domain randomization \cite{domainrandom} and its performance should be quantitatively evaluated.

    In the current work, the language instruction task was limited to 2D tasks such as resource collection, with training conducted by randomly sampling instruction vectors without using an LLM. To extend the method to a broader range of tasks, further improvements are needed in converting natural language into numerical representations.

    \section{CONCLUSION}
   
    This paper proposed ICCO, a MARL framework that balances language-instruction following and cooperative task execution. ICCO consists of a Coordinator, which generates TACI from language inputs and global observations. The coordination and local policies are consistently trained within the CTDE framework to balance instruction following and task requirements.
    Experiments show that ICCO outperforms baseline methods in balancing instruction following and task execution. It also effectively handles natural language instructions with an LLM. Notably, the Coordinator cannot be replaced by an LLM with prompt engineering, highlighting the limitations of relying solely on prompts for task execution.
    As future work, we aim to develop a general approach for converting natural language into numerical representations and to evaluate this approach on more challenging tasks, including cooperative manipulation.
   
\begin{figure}[t]
        \begin{center}                
        \includegraphics[width=0.41\textwidth]{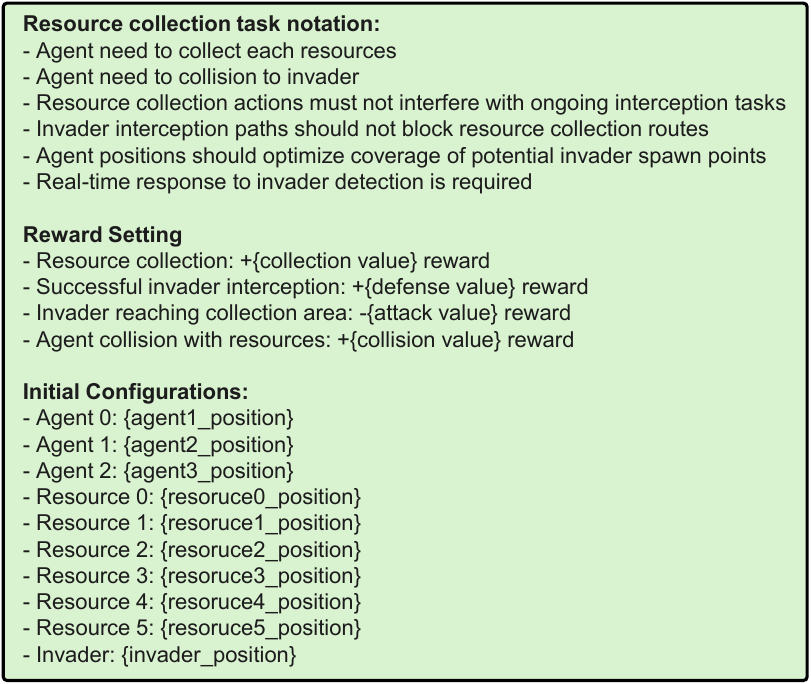}
        \vspace{-9pt}
            \caption{Key components of the task-aligned prompt: Elements enclosed in \{\} represent variables that are updated when inputted to the LLM. All resource and invader rows provide position information.}
            \label{fig:prompt2}
        \end{center}
    \end{figure}

    \appendix
    This appendix describes the prompt used in Subsection \ref{sec: ex_coordinator_policy}. Since some elements overlap with those in Fig. \ref{fig:prompt}, Fig. \ref{fig:prompt2} highlights the additional components for the task-aligned prompt. The task-aligned prompt includes resource collection and invader defense as task details, rewards identical to those used in ICCO, and the initial configurations that specify the positions of all agents, resources, and the invader. 
    
\bibliographystyle{IEEEtran}
\bibliography{IEEEexample}             % bib file to produce the bibliography

\end{document}